\pdfoutput=1
\documentclass[11pt]{article}

\usepackage{acl}

\usepackage{times}
\usepackage{url}
\usepackage{latexsym}

\usepackage{multicol}
\usepackage{colortbl}
\usepackage{multirow}
\usepackage{array}

\usepackage{caption}

\usepackage{enumitem}
\usepackage{soul}
\usepackage{arydshln}
\usepackage{booktabs}

\usepackage{graphicx}

\usepackage{microtype}
\usepackage{inconsolata}
\usepackage{bm}

\usepackage[T1]{fontenc}
\usepackage[utf8]{inputenc}
\usepackage{adjustbox}
\usepackage{arydshln}
\usepackage{amssymb}% http://ctan.org/pkg/amssymb
\usepackage{pifont}% http://ctan.org/pkg/pifont
\usepackage{nicematrix}
\usepackage{siunitx} 
\usepackage{titletoc}
\usepackage[disable]{todonotes}  % use option "disable" to suppress notes
\makeatletter
\newcommand*\iftodonotes{\if@todonotes@disabled\expandafter\@secondoftwo\else\expandafter\@firstoftwo\fi}  % defines \iftodonotes{<true>}{<false>}, thanks to https://tex.stackexchange.com/questions/126559/conditional-based-on-packageoption
\makeatother
\newcommand{\noindentaftertodo}{\iftodonotes{\noindent}{}\ignorespaces}
\newcommand{\note}[4][]{{\todo[author=#2,color=#3,size=\scriptsize,fancyline,caption={},#1]{#4}}} % default note settings, used by macros below.
\newcommand{\response}[1]{\vspace{3pt}\hrule\vspace{3pt}\textbf{#1:}} % insert \response{myname} within someone else's todonote
\newcommand{\jason}[2][]{\note[#1]{Jason}{orange!40}{#2}}
\newcommand{\boshi}[2][]{\note[#1]{Boshi}{purple!40}{#2}}

\newcommand{\Jason}[2][]{\jason[inline,#1]{#2}\noindentaftertodo}

\newcommand{\marktext}[2]{\adjustbox{bgcolor=#1}{\strut #2}}
\newcommand{\myblue}[1]{\marktext{blue!30}{\texttt{#1}}}
\newcommand{\mypink}[1]{\marktext{pink!70}{\texttt{#1}}}

\newcommand{\nop}[1]{}

\newcommand\ttsmall[1]{\texttt{\textmd {#1}}}

\title{LLMs in the Imaginarium:\\ Tool Learning through Simulated Trial and Error}

\author{\
Boshi Wang$^\text{\ding{171}}$\thanks{Work done as an intern at Microsoft Semantic Machines.}%
\ \ \ 
Hao Fang$^\Diamond$%
\ \ \ 
Jason Eisner$^\Diamond$%
\ \ \ 
\textbf{Benjamin Van Durme$^\Diamond$}%
\ \ \ 
\textbf{Yu Su$^\Diamond$}%
\\$^\text{\ding{171}}$The Ohio State University \hspace{1cm} $^\Diamond$Microsoft Semantic Machines \\ 
{\small \texttt{wang.13930@osu.edu},
\texttt{\{hao.fang,jason.eisner,bevandur,yusu2\}@microsoft.com}} \\
}

\begin{document}
\maketitle

\begin{abstract}
Tools are essential for large language models (LLMs) to acquire up-to-date information and take consequential actions in external environments.
Existing work on tool-augmented LLMs primarily focuses on the broad coverage of tools and the flexibility of adding new tools.
However, a critical aspect that has surprisingly been understudied is simply \textit{how accurately an LLM uses tools for which it has been trained}.
We find that existing LLMs, including GPT-4 and open-source LLMs specifically fine-tuned for tool use, only reach a correctness rate in the range of \num{30}\% to \num{60}\%, far from reliable use in practice.
We propose a biologically inspired method for tool-augmented LLMs, simulated trial and error (STE), that orchestrates three key mechanisms for successful tool use behaviors in the biological system: trial and error, imagination, and memory.
Specifically, STE leverages an LLM's `imagination' to simulate plausible scenarios for using a tool, after which the LLM interacts with the tool to learn from its execution feedback. 
Both short-term and long-term memory are employed to improve the depth and breadth of the exploration, respectively.
Comprehensive experiments on ToolBench show that STE substantially improves tool learning for LLMs under both in-context learning and fine-tuning settings, bringing a boost of \num{46.7}\% to Mistral-Instruct-7B and enabling it to outperform GPT-4. 
We also show effective continual learning of tools via a simple experience replay strategy.\footnote{Code and data available at \url{https://github.com/microsoft/simulated-trial-and-error}.} 

% Augmenting large language models (LLMs) with external tools can compensate for many of their limitations as general problem-solving agents, such as fixed knowledge storage and limited computational capacity.  Existing work mostly focuses on developing pipelines that enable flexible adaptation to unseen tools.  Thus, improving the usage of known tools remains an understudied but important bottleneck. Motivated by how humans typically master tools through active engagement and reinforcement, we propose a tool learning pipeline that consists of two stages: 1) an exploration stage where the LLM extensively interacts with the tool, with two memory mechanisms designed to dynamically grow trial and error experiences which enable progressive tool learning over a horizon; 2) an exploitation stage where the gathered experiences are distilled into a set of tool-use examples that are utilized to enhance LLMs' tool-use ability. Experiments on ToolBench show that the proposed pipeline greatly improves LLMs' tool-use performance, and that fine-tuning largely outperforms in-context learning by injecting a much wider range of experiences into the model. While vanilla fine-tuning suffers from catastrophic forgetting, we show that a simple rehearsal strategy could largely mitigate this issue, allowing the model to continually learn new tools while preserving its previously acquired skills.
\end{abstract}

\section{Introduction}
\label{sec:intro}

% \begin{itemize}
%     \item Why tools are important for LLMs, recent development
%     \item Limitations of existing efforts
%     \item Simulated trial and error, simulation, environmental feedback
%     \item The three desiderata achieved through our method
%     \item Highlights of results
% \end{itemize}

Tools play an essential role in extending humans~\citep{gibson1993tools} and other animals~\citep{shumaker2011animal} beyond the confines of physical bodies to better perceive and exert impact on their environment.   
There is a recent surge of interest in augmenting large language models (LLMs) with tools to transcend the confines of their static parametric knowledge and text-in-text-out interface, empowering them to acquire up-to-date information, call upon external reasoners, and take consequential actions in external environments~\citep{schick2023toolformer,mialon2023augmented,qin2023tool}.\jason{It may be a good idea to explain what a tool is---it's just an API and the LLM has to generate a textual function call licensed by the API, which is called to return a string that is appended to the context so that the LLM can subsequently condition on it.}

Existing work on tool-augmented LLMs primarily aims to increase the ease of adding new tools, or the ability to access many tools (e.g., up to \num{16000} APIs~\citep{qin2023toolllm}).
This is achieved through one of two common approaches: 1) In-context learning (ICL), which prompts frozen LLMs with API specification and tool use examples (i.e., instruction-API call pairs)~\citep{lu2023chameleon, song2023restgpt, shen2023hugginggpt, liang2023taskmatrix}, or 2) fine-tuning with tool use examples synthesized by LLMs~\citep{schick2023toolformer, patil2023gorilla, qin2023toolllm, tang2023toolalpaca}.
While coverage and flexibility are important for tool use, a critical aspect that, perhaps surprisingly, has been understudied is simply \textit{how accurately an LLM uses tools for which it has been trained}.
ICL is flexible but hard to drive to production-level accuracy.
Fine-tuning can potentially lead to better accuracy by integrating a larger number of examples, but existing work mostly focuses on generalizing to unseen tools instead of optimizing an LLM's ability to use tools seen during training~\citep{qin2023toolllm, patil2023gorilla, tang2023toolalpaca}. 
Meanwhile, practical deployment of tool-augmented LLMs necessitates a high level of accuracy as they enable consequential actions, e.g., financial transactions or other legally binding operations.
Inaccurate tool use could lead to undesired or harmful outcomes and quickly undermine user trust.

\begin{figure*}[t]
  \centering
    \includegraphics[width=0.95\linewidth]{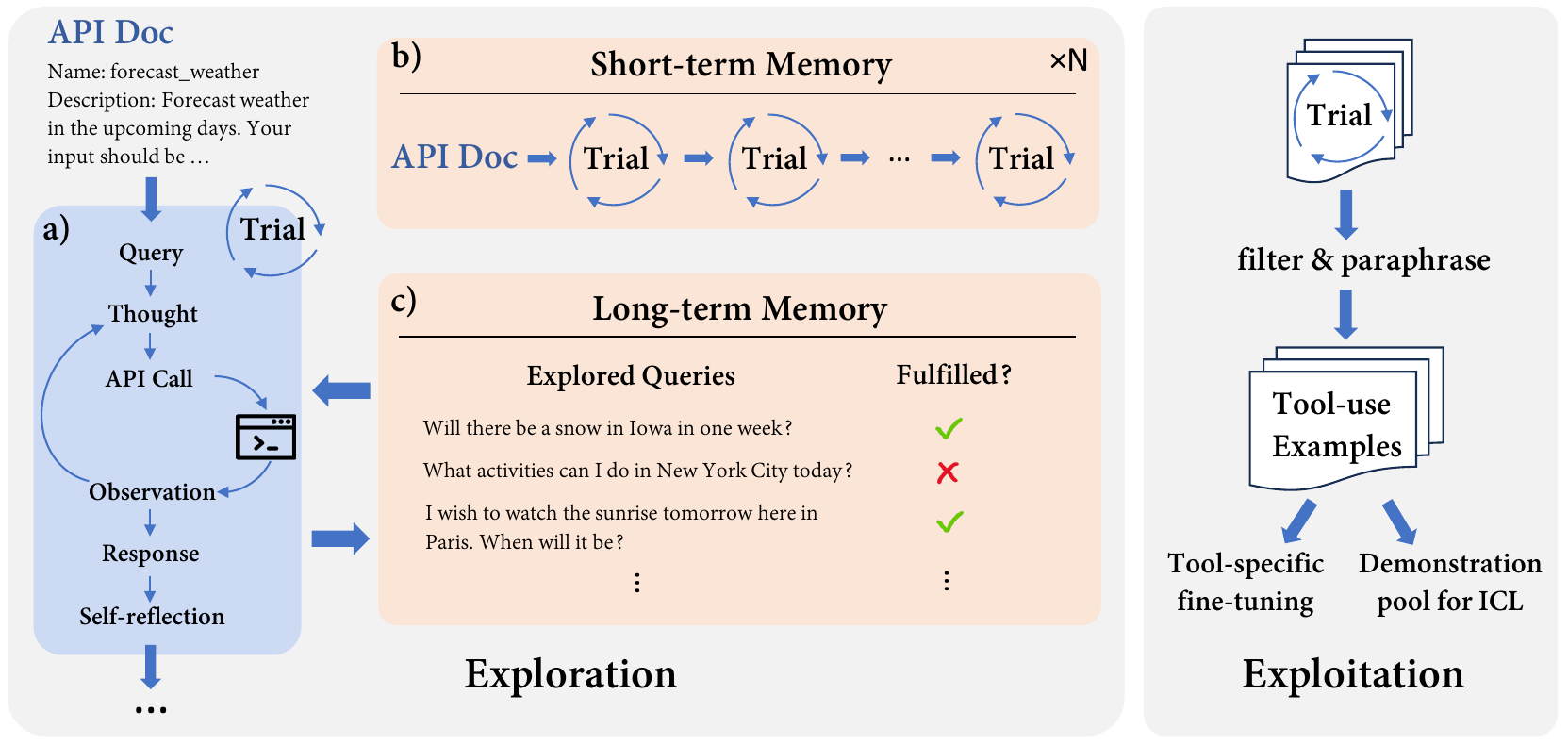}
    \vspace{-5pt}
  \caption{Illustration of simulated trial and error. In the exploration stage, an LLM interacts with the tool and progressively gathers tool-use experiences through trial and error. Specifically, a) in each trial, the LLM imagines plausible scenarios related to the target tool, iteratively interacts with the tool to fulfill the user query, and in the end self-reflects on the trial; b) a short-term memory consisting of recent trial trajectories encourages learning from fine-grained successes and failures and exploring the API in greater depth; c) a long-term memory of coarse-grained past trial and error experiences maintains progressive learning over a long time horizon. In the exploitation stage, the exploration experiences are distilled into a set of tool-use examples for either ICL or fine-tuning.}
\label{fig1}
\vspace{-10pt}
\end{figure*}

\textit{How to truly master a tool?} 
We turn to successful precedents in the biological system such as humans~\citep{gibson1993tools}, apes and corvids~\citep{emery2004mentality}.
Learning to use a tool is a rather advanced cognitive function that depends on many other cognitive functions.
First of all, \textit{trial and error} is essential for tool learning~\citep{beck1973observation,auersperg2011flexibility}. 
We do not master a tool solely by reading the `user manual'; rather, we explore different ways of using the tool, observe the outcome, and learn from both successes and failures.
Furthermore, intelligent animals do not just do random trial and error---we proactively \textit{imagine} or \textit{simulate} plausible scenarios that are not currently available to perception for exploration~\citep{emery2004mentality,redish2016vicarious}.  
Finally, \textit{memory}, both short-term and long-term, is instrumental for the progressive learning and recurrent use of tools~\citep{vaesen2012cognitive,emery2004mentality,clayton1998episodic}.

To this end, we propose \textit{simulated trial and error} (STE; illustrated in Figure~\ref{fig1}), a biologically inspired method for tool-augmented LLMs.
Given a tool (e.g., an API with its specification), STE leverages an LLM to simulate, or `imagine', plausible scenarios (i.e., instructions) for using the tool. It then iteratively interacts with the API to fulfill the scenario by synthesizing, executing, and observing the feedback from API calls, and then reflects on the current trial~\citep{shinn2023reflexion}. 
We devise memory mechanisms to improve the quality of the simulated instructions. 
A short-term memory consisting of recent trial and error trajectories is employed to facilitate deeper exploration in a single episode, while a long-term memory containing distilled past exploration and reflections maintains progressive learning over a long horizon.
In the exploitation stage, one can use the tool use examples from the explored trials to fine-tune an LLM, or simply do ICL by retrieving from those examples.

We conduct comprehensive experiments on APIs from ToolBench~\cite{qin2023toolllm} and summarize the main findings as follows:

\begin{itemize}[leftmargin=12pt,itemsep=2pt,topsep=0pt,parsep=0pt,partopsep=0pt]
\item Existing LLMs are far from reaching reliable tool use performance: GPT-4~\cite{openai2023gpt4} gets \num{60.8}\% correctness, and ToolLLaMA-v2~\cite{qin2023toolllm} that was specifically fine-tuned for tool use only gets \num{37.3}\%. 

\item STE proves to be remarkably effective for augmenting LLMs with tools, under both ICL and fine-tuning settings. STE improves the tool use capability of Mistral-Instruct-7B~\cite{jiang2023mistral} to \num{76.8}\% (a boost of \num{46.7}\% absolute), making it outperform GPT-4 with ICL.

\begin{figure*}[ht]
  \centering
    \includegraphics[width=1.0\linewidth]{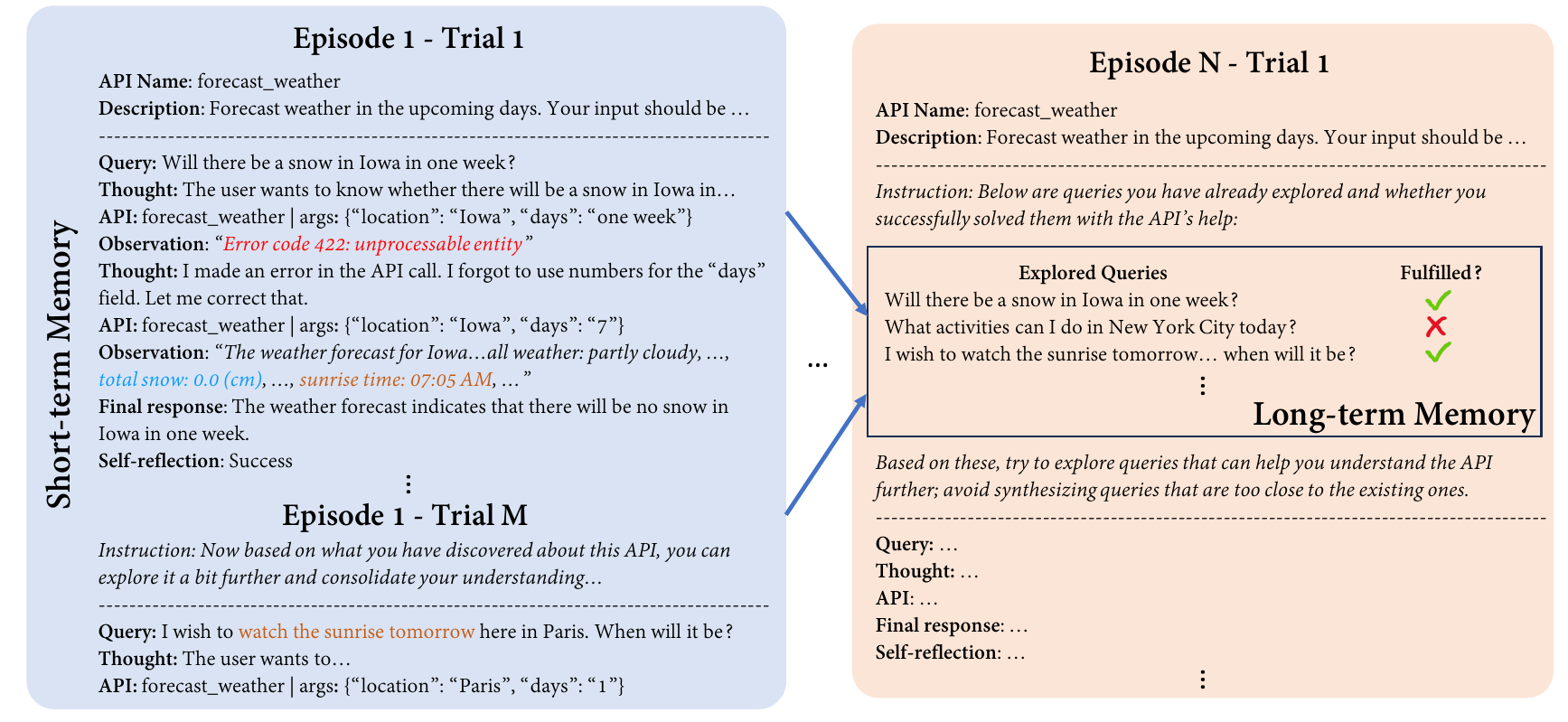}
  \caption{Exploration with simulated trial and error, highlighting the memory mechanisms. Each episode begins with the API specification (only in the first trial), followed by a series of trials dynamically added in the short-term memory. The long-term memory is loaded into the context at the beginning of every trial to allow the LLM to progressively imagine novel scenarios, and then offloaded afterward (omitted in the figure).}\Jason{A few suggestions to improve the figure, which has great content but is a bit confusing when I try to understand the prompt format.  (1) I think the API name and description only appear at the start of each episode and are not repeated at the start of each trial.  So shouldn't they appear before the label "Trial 1"?   Or perhaps they appear at the start of Trial 1 only, and the instruction "Now based on what you have discovered ..." appears at the start of all trials $> 1$?  This isn't clear from the figure itself; so you can explain in the caption if necessary.
  (2) Isn't this also true of Long-Term Memory---it is not repeated at the start of each trial? (The main text says that it is, but surely that would be a waste of prompt space \ldots ?) (3) The bottom part of the right panel is short-term memory again, right?  So can you label it as such?  This makes it clear that the right side is not showing LTM, as the reader might otherwise think, but is showing Episode N with both LTM and STM.  I suggest using a bracket in the left panel to delineate the "Query" through "Self-reflection" portion as a short-term memory.  You can then use the same kind of bracket in the right panel.  The caption should clarify that LTM is omitted in episode 1 but appears at the start of all subsequent episodes---is that right?}
\label{fig:mem}
\end{figure*}

\item In practice, new tools are continually added. Fine-tuning (with or without STE) brings the challenge of catastrophic forgetting, where learning new tools may cause the LLM 
% may fail to retain 
to lose its existing tool use capabilities or general language capabilities.
We demonstrate that a simple experience replay strategy~\citep{scialom-etal-2022-fine} could largely mitigate the issue, allowing the model to continually learn new tools while preserving its previously acquired skills.
\end{itemize}

% We conduct experiments on ToolBench\footnote{https://github.com/OpenBMB/ToolBench}, and show that our proposed framework could greatly enhance the LLM's tool-use performance, and also that fine-tuning is much more effective than ICL in equipping LLMs with tools, presumably because fine-tuning allows distilling a much larger set of experiences into the model. While vanilla fine-tuning would cause catastrophic forgetting which limits the flexibility aspect, we show that a simple experience replay strategy~\citep{scialom-etal-2022-fine} could largely mitigate the issue, allowing the model to continually learn new tools while preserving its previously acquired skills.

\section{Simulated Trial and Error}
We introduce our proposed simulated trial and error (STE) for tool learning.
STE consists of an exploration phase and an exploitation phase,\jason{Why not "learning" and "deployment"?  There is nothing fancy about the exploitation phase (except maybe fine-tuning); you're just training on the collected examples.  You're not shifting gradually from exploration to exploitation as in most RL.}\boshi{Nice point! Not sure about "deployment" though. Do you see this as important to change here?} which are discussed next.

\subsection{Exploration}
In the exploration phase, for each new API, the LLM interacts with the API within a budget in order to gain as much information as possible about the API. The exploration phase consists of a series of trials (Figure~\ref{fig1}) resembling humans' progressive learning of a tool. In each trial, conditioned on the API description, the LLM 1) imagines a plausible user query relevant to the API; 2) tries to fulfill the query by interacting with the API; 3) reflects on the trial to facilitate subsequent exploration. Three core design components are integrated with the trials to enhance the validity, comprehensiveness and diversity of the exploration, introduced next.

\noindent\textbf{Iterative self-refine with execution feedback.} To improve the validity of the exploration, we use a strategy similar to the ideas of \citet{chen2023teaching, qin2023toolllm, madaan2023self, shinn2023reflexion} where the LLM learns from the execution feedback to refine its own outputs (Figure \ref{fig:mem}, top left). Specifically, we adopt the ReAct format~\cite{yao2022react} where during each step, the LLM first verbalizes its internal thought, then makes an action (API call) and observes the corresponding execution feedback, and then repeats the thought$\rightarrow$action$\rightarrow$observation process until the model decides that the API call has returned adequate information or a predefined maximum number of calls. During this stage, the LLM learns from the execution environment to correct its own syntactic and semantic errors in API calls, gathering tool-use experiences as fine-grained trial-and-error trajectories. Afterward, the model responds to the user's query and self-reflects on whether the explored query is successfully fulfilled or not.\jason{what enforces the ReAct format---instructions? few-shot examples?  constrained decoding?}\boshi{just instructions; there's also a parser which gives error msg if the format doesn't pass}

\noindent\textbf{Short-term memory.} A direct implementation of the exploration where each trial is conducted in a separate episode only allows shallow explorations of the API. We augment the LLM with a short-term memory consisting of the exploration trajectories of recent trials, where the LLM is instructed to conduct subsequent trials conditioned on the memory (Figure \ref{fig:mem}, left). Each episode starts with a fresh short-term memory, where newly conducted trials are dynamically added into the memory for a certain number of trials. This allows the model to learn from recent fine-grained successes and failures (e.g., syntax/semantic errors), and also explore the API in greater depth in the coming trials based on its previous observations of the API (e.g., unexpected functionalities).

\noindent\textbf{Long-term memory.} Only a small number of trials can be stored in short-term memory since the fine-grained trajectories quickly consume the LLM's context capacity.
% \jason{Would LTM be unnecessary if the context window (STM) was unbounded?  Should we think of LTM as just a compressed version of what you'd put in the unbounded memory if you had one?  Worth discussing.\response{\boshi}Right really nice point. Added some discussion \response{jason} Looks good to me!} 
We augment the LLM with a long-term memory that stores distilled trial-and-error experiences from past episodes, in order to support progressive learning over a long time horizon.
Specifically, the long-term memory records the past-explored queries and whether they were judged as successfully fulfilled (Figure \ref{fig:mem}, right).
It is only loaded into the context at the beginning of every new trial, where the model is instructed to imagine scenarios that are distant from previously explored ones to improve information gain.
In this way, the long-term memory serves as a growing pool of past successes and failures, which allows the LLM to continually expand the exploration in order to make progress across different episodes.\jason{Should acknowledge that we are trusting the LLM to sample wisely, as it was instructed to do.  Do we evaluate whether it actually manages to do so?}\boshi{I changed the wording a bit to say it allows the LLM to XXX. Hard to evaluate this but we have a case study in later sections which shows it}  

\subsection{Exploitation}
In the exploitation stage, the trials obtained from the exploration stage are utilized to enhance the tool-use ability of an LLM via either fine-tuning or in-context learning (ICL).
For each trial, we extract the synthesized user query, the LLM's last API call and its execution results, and the final response from the trial trajectory. Then, we perform filtering by using GPT-4 to judge the validity of each example, and then paraphrase the valid examples\jason{are these the same as what you called "successful queries" in the previous subsection? \response{boshi} not exactly but close; here the filtering is more rigorous (e.g., groundedness of response w.r.t. the API results, etc.) which GPT-4 does better. We could also use GPT-4 instead of ChatGPT during exploration but it was not done due to various reasons (didn't have GPT-4 budget initially, etc.) \response{jason} The issue is that you say "valid" without defining it.  Can this paragraph point to somewhere where it is defined?  You may have to add this in the appendix: I can't find any relevant future mentions of the word "valid"} for each new API into approximately the same amount (Appendix~\ref{sec:prompts}), which maintains a balance across different APIs and further adds linguistic variations into the synthesized tool-use examples.

For fine-tuning, we use the standard language modeling objective where the loss is computed only for the tool-use/response generation part, and do not include the API documentation in the context.
For ICL, the synthesized examples are used as the demonstration pool from which in-context examples are retrieved and appended to the API documentations in the LLM's context.
We use a dynamic nearest-neighbor demonstration selection strategy where the examples that are semantically closest to the test user query are retrieved as in-context examples, one of the top performing strategies for ICL~\cite{liu-etal-2022-makes, rubin-etal-2022-learning}.
\section{Experimental Setup}
\label{sec:exp_setup}

\noindent\textbf{Tools.}
We conduct experiments using APIs from ToolBench~\cite{qin2023toolllm}, a large-scale repository of real-world APIs collected from RapidAPI and BMTools.
We filter down to the APIs that are free to use with low execution latency.
In the end, we obtain \num{50} APIs that span search engines (e.g., Google Search \& Places), domain-specific information-seeking APIs (e.g., Wikipedia, Weather, Sports, Gaming), and also problem-solving ones such as WolframAlpha, Number Translator, etc. More details are in Appendix~\ref{app:api}.

\begin{table*}[!htbp]
\centering
\small

\resizebox{0.85\linewidth}{!}{%
\begin{tabular}{llccccc}
\toprule

\textbf{Setting} & \textbf{Base Model}  & \textbf{Wellformed?} & \textbf{API Match} & \textbf{Correctness} \\
\toprule

\multirow{6}{*}{Baseline} & ToolLLaMA-v2  & \underline{\num{98.1}} & \num{49.0} & \num{37.3} &	 \\
% \hdashline\noalign{\vskip 0.5ex}
\cdashline{2-7}  \noalign{\vskip 0.5ex}

& Llama-2-Chat-7B & \num{34.5} &	\num{40.2}& \num{10.7} &	 \\
                            
                            & Llama-2-Chat-13B & \num{79.3}&	\num{53.6}& \num{32.7} &	 \\
                            & Mistral-Instruct-7B & \num{61.7}&	\num{69.3}& \num{30.1} &	 \\

\cdashline{2-7}  \noalign{\vskip 0.5ex}
                            
                            & GPT-3.5-turbo (16k-0613) & \num{96.9}&	\num{77.6}& \num{60.5} &	 \\
                            & GPT-4 (8k-0613) & \num{96.1} &	\underline{\num{78.1}}& \underline{\num{60.8}} &	 \\
\midrule

\multirow{5}{*}{\cellcolor{white}ICL w/ STE} 
                            & Llama-2-Chat-7B & \num{58.3}&	\num{86.7}&	\num{41.7}\\
                            & Llama-2-Chat-13B & \num{87.5} &	\num{86.6}&	\num{62.9}\\
                            & Mistral-Instruct-7B & \num{69.9}&	\num{88.4}&	\num{47.9}\\
\cdashline{2-7}  \noalign{\vskip 0.5ex}
                            & GPT-3.5-turbo (16k-0613) & \num{97.6}&	\num{90.8}&	\num{75.6}\\
                            & GPT-4 (8k-0613) & \underline{\num{97.7}}&	\underline{\num{92.8}}&	\underline{\num{76.3}}\\
                                
%\hdashline\noalign{\vskip 0.5ex}
\midrule

\multirow{3}{*}{\cellcolor{white}Fine-tuning w/ STE} 
                            & Llama-2-Chat-7B & \underline{\textbf{99.2}} &	\num{94.9} & \num{73.3} \\
                            & Llama-2-Chat-13B & \num{98.9}&	\num{95.1}& \num{74.3} \\
                            & Mistral-Instruct-7B &\num{99.1}&	\underline{\textbf{95.8}}& \underline{$\textbf{76.8}$} \\
\bottomrule
\end{tabular}%
}
\caption{Overall tool-use performance. STE is effective when used in both ICL and fine-tuning. Best overall results are \textbf{bold-faced}, and best results under each setting are \underline{underscored}.}
\label{tbl:main}
\vspace{-10pt}
\end{table*}
\begin{table}[t]
\centering
\small
\resizebox{1.0\linewidth}{!}{%
\begin{tabular}{lcccc}
\toprule

\textbf{Setting}  & \textbf{Wellformed?} & \textbf{API Match} & \textbf{Correctness} \\
\toprule

Full STE  & \num{99.2} &	\num{94.9} & \num{73.3}&	 \\
\;\; -- Exec.\ feedback & \num{89.9} &	\num{79.4} & \num{50.5} \\
\;\; -- Short.\ Mem.\ &\num{99.7}&\num{70.6} &\num{53.9}\\
\;\; -- Long.\ Mem.\ & \num{98.7}&	\num{79.9}&	\num{59.7}\\
\;\; -- Self-reflection &\num{99.3}&\num{81.7}& \num{60.1}\\

\bottomrule
\end{tabular}%
}
\caption{Results for ablations. We separately ablate each key component of our exploration design. Exploitation is done by fine-tuning Llama-2-Chat-7B.}
\label{tbl:abl}
\vspace{-10pt}
\end{table}

\noindent\textbf{Setup for exploration.}
In the exploration stage, we use ChatGPT (\ttsmall{16k-0613}) for exploration and paraphrasing, and GPT-4 (\ttsmall{8k-0613})~\citep{openai2023gpt4} for final example filtering.
We set the maximum number of API calls for each trial to be \num{4}.
For each API, the exploration stage lasts for \num{15} episodes with \num{4} trials per episode, resulting in a total of 60 examples before filtering and paraphrasing.
After filtering, \num{15} examples for each API are randomly selected into the test set, where the remaining ones are paraphrased into $\sim$\num{140} examples, making a total of $\sim$\num{7}K tool-use examples.
For the test examples, we manually examine and correct any issues, if any, to ensure test set quality.

\noindent\textbf{Baselines \& exploitation with STE.}
We experiment with \textbf{Llama-2-Chat-7B/13B}~\cite{touvron2023llama}, \textbf{Mistral-Instruct-7B}~\cite{jiang2023mistral}, and \textbf{GPT-3.5-turbo}/\textbf{GPT-4} (ICL only) and compare their performance with and without STE. We compare with \textbf{ToolLLaMA-v2}~\citep{qin2023toolllm} as the main baseline for existing tool learning strategies. 
It is based on Llama-2-7B and fine-tuned on \num{126}K tool use examples synthesized by ChatGPT-3.5-turbo for general tool-use, covering a large number of tools from RapidAPI including the ones used in our experiments.  

For ICL with nearest neighbor demonstration selection, following prior work~\cite{liu-etal-2022-makes, rubin-etal-2022-learning}, we use the \ttsmall{paraphrase-mpnet-base-v2} model from SentenceBERT~\citep{reimers-gurevych-2019-sentence} for computing the semantic similarity, and choose the top \num{8} examples closest to the test query as in-context demonstrations.
For Llama-2 with ICL, since the token length of the full \num{50} API documentations (around \num{7}K tokens) is beyond its context length (\num{4096}),\footnote{While there exist variants of Llama-2 with longer context (e.g., \citet{xiong2023effective}), we stick to the original model in \citet{touvron2023llama} for fair comparison.}
we augment the model with an oracle tool retriever which retrieves the top \num{15} similar APIs \textit{w.r.t.}\ the ground truth API using the associated documentation. We augment other models of similar scales (\num{7}B/\num{13}B) with the same tool retriever (when ICL is used for exploitation) for fair comparison.
LLMs fine-tuned on STE do not need such API documentation in the context, which substantially reduces the inference cost.

\noindent\textbf{Evaluation metrics.}
We evaluate the model by matching the predicted API call against the ground truth.
For APIs that have strict value ranges for the arguments, we perform string matching on the respective arguments directly.
For APIs that accept free natural language inputs, we use ChatGPT to judge the correctness of the predicted API arguments.
We report the overall accuracy considering both API name and arguments (\textbf{Correctness}) as the main metric, together with the percentage of examples with valid API calls and no syntax/formatting errors (\textbf{Wellformedness}) and the percentage of examples that correctly choose to use the ground-truth API (\textbf{API Match}).
While it is desirable to also evaluate the model regarding whether the model resolves the user query successfully based on the execution results, the majority of the APIs in our experiments are connected to dynamic real-world information (e.g., weather `tomorrow' where the date is contingent on the actual time of making the API call), which makes such evaluation infeasible.
We leave this challenge to future research.
\section{Results}
\subsection{Effectiveness of STE}
Results are included in Table~\ref{tbl:main}. We summarize the main findings below.

\noindent\textbf{None of the baseline models displays satisfactory performance.} For all of the baseline models that we tested, none of them achieves a satisfactory tool-use performance. The best model is GPT-4, which only achieves an overall correctness rate of \num{60.8}\%. Llama-2 and Mistral achieve a much lower performance, largely due to the model not being able to follow the specified syntactic/formatting requirements when making API calls.\footnote{The superior ability of GPT-3.5/GPT-4 to follow the syntax may be partially due to their special enhancement on function-calling (https://openai.com/blog/function-calling-and-other-api-updates).} ToolLLaMA-v2~\citep{qin2023toolllm}, despite extensively fine-tuned for tool use, still largely underperforms GPT-3.5/4. 
Its performance improvement over non-fine-tuned baselines like Llama-2 seems to mainly come from wellformedness, and it still faces severe difficulties in choosing the correct tool and predicting the right arguments. This suggests that fine-tuning for general tool use is insufficient for achieving the level of performance needed for practical deployment.

\noindent\textbf{STE is effective with both ICL and fine-tuning.}
Remarkable gains are observed under both settings.
When retrieving from the tool use examples generated by STE for ICL, we see improvements across the board, with up to \num{30.2}\% (for Llama-2-Chat-13B) in correctness for open-source LLMs.
It also boosts the already strong performance of GPT-3.5/4 substantially.
Fine-tuning with STE examples improves the tool use capability of open-source LLMs by an even larger margin, boosting Mistral-Instruct-7B by \num{46.7}\% in correctness and enabling it to outperform GPT-4.
Fine-tuning with STE also makes LLMs almost perfect in wellformedness and choosing the right tools.
This is likely because fine-tuning allows injecting a much wider range of tool use examples into a model than ICL.
While we cannot fine-tune GPT-3.5/4 due to cost and availability, it is plausible to hypothesize that STE could further improve their tool-use ability beyond their current ICL performance.

\subsection{Ablation Studies}
\label{sec:analysis}
We conduct an ablation study for our exploration design, with exploitation done by fine-tuning Llama-2-Chat-7B. 
Specifically, we ablate the execution feedback, short/long-term memory, and the self-reflection component. We extend the number of episodes to preserve the total number of trials if needed. The results in Table~\ref{tbl:abl} show that 1) exploration without execution feedback could give a notable amount of ill-formed examples where the API calls do not follow the syntax/formatting requirements; 2) both short-term and long-term memory prove to be essential for effective trial and error; 3) self-reflection is important in maintaining an informative long-term memory for exploration. To better understand the benefits of our memory design, we conduct a case study with the \ttsmall{forecast\_weather} API (examples in Appendix~\ref{app:examples}), which clearly shows that both of the memory mechanisms substantially improve the diversity and comprehensiveness of exploration:

\begin{itemize}[noitemsep,nolistsep,leftmargin=*]
    \item\textbf{Short-term memory boosts specificity and comprehensiveness.} Comparing the trials with/without short-term memory, it can be seen that the short-term memory effectively drives the LLM to comprehensively explore fine-grained information from the tool, spanning \num{16} different attributes (e.g., humidity, precipitation, UV index, visibility, and sunrise/sunset time) in total. Meanwhile, when short-term memory is disabled, the examples are much less specific and mostly about general weather conditions (e.g., \textit{``What will be the weather like...''}) due to the inability to leverage newly obtained information from the execution results. In addition, exploration without short-term memory results in a significantly lower percentage of positive tool-use examples (\num{78.3}\% $\rightarrow$ \num{51.7}\%), since the model cannot learn from fine-grained past errors to facilitate future trials. As an example, the model synthesizes a considerable amount of queries where the time is specified as the day of the week, which is not a supported parameter type of the API and hence constantly results in failures. 
    
    \item\textbf{Long-term memory improves overall diversity over a long time horizon.} With long-term memory, the LLM explores examples covering a broader range of subjects, and maintains the progress over different episodes. When long-term memory is disabled, the trials across episodes become repetitive and less informative. For quantitative characterization, we extract the core subjects (location, time, attribute) from the queries, measure their diversity and also plot the attribute distribution (Appendix~\ref{app:examples}). With long-term memory, all the queries are distinct and the trials are balanced across different attributes. Without long-term memory, only \num{71.7}\% of the trials concern distinct subjects and the distribution across attributes is much more skewed, showing the effectiveness of long-term memory in maintaining the diversity of exploration over a long time horizon.
    
\end{itemize}

\begin{figure*}[!htbp]
  \centering
    \includegraphics[width=0.95\linewidth]{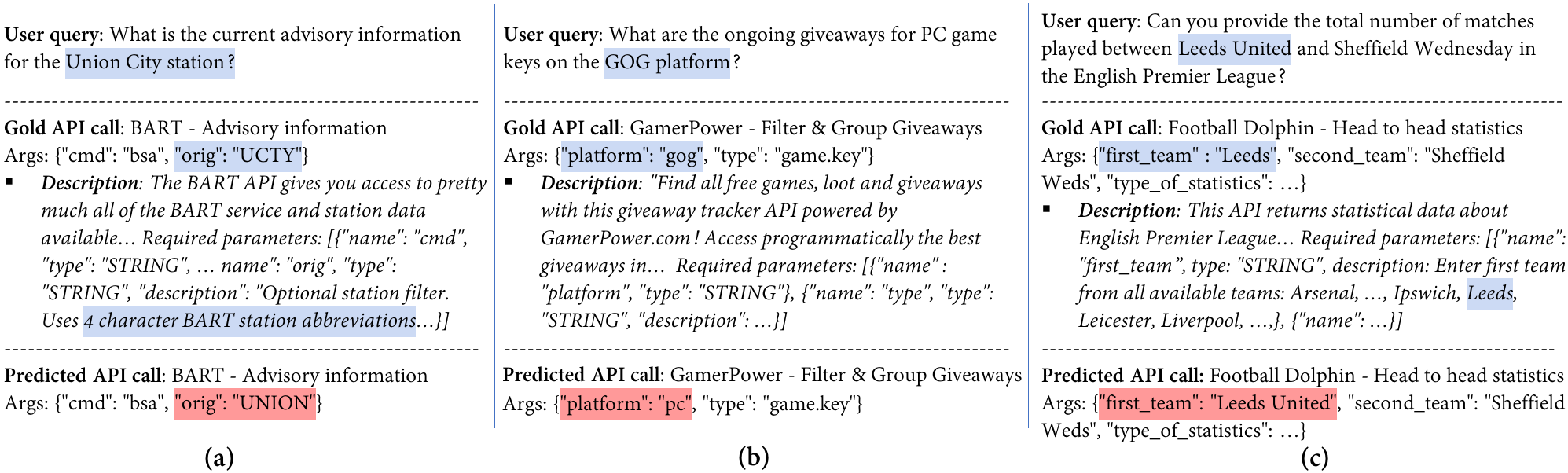}
    \vspace{-8pt}
  \caption{Error examples of Mistral-Instruct-7B after fine-tuning: (a) commonsense/world knowledge, (b) language understanding, and (c) grounding.}
\label{fig:mistral}
\end{figure*}

\begin{table*}[!htbp]
\centering
\small
\resizebox{0.9\linewidth}{!}{%
\begin{tabular}{lcccccccc}
\toprule

& \textbf{Batch 1}  & \textbf{Batch 2} & \textbf{Batch 3} &  \textbf{Batch 4} & \textbf{All APIs} & \textbf{MMLU} & \textbf{BBH}\\

\midrule
Llama-Flan & - & - &	-& - & - & \num{37.2} & \num{39.5}\\
\midrule

CL-Round 1 & \num{80.6} & - &	-& - & - & \num{39.6} & \num{36.8}\\

CL-Round 2 & \num{1.7} $\rightarrow$ \num{76.1} & \num{87.7} $\rightarrow$ \num{84.1} &	-& - &  - & \num{40.2} & \num{38.9}\\

CL-Round 3 & \num{0.0} $\rightarrow$ \num{70.6} & \num{56.9} $\rightarrow$ \num{84.1} &	\num{68.9} $\rightarrow$ \num{65.6}& - & - & \num{39.2} & \num{37.5}\\

CL-Round 4 & \num{0.0} $\rightarrow$ \num{65.0} & \num{38.5} $\rightarrow$ \num{88.7} &	\num{25.0} $\rightarrow$ \num{66.1}& \num{71.8}$\rightarrow$ \num{70.3}& \num{34.7} $\rightarrow$ \num{72.8} & \num{38.5} & \num{39.1}\\

\midrule

Llama-FT & \num{73.3} & \num{87.2} &	\num{68.3}& \num{67.2} & \num{74.1} & \num{38.7} & \num{40.8}\\

\bottomrule
\end{tabular}%
}
\caption{Results for continual learning. Llama-Flan is the base LLM and Llama-FT is Llama-Flan fine-tuned on all the tools at once. For CL, the tools are split into four batches and the LLM needs to continually learn a new batch in each round. Scores to the left/right of each arrow (``$\rightarrow$'') are the tool-use correctness without/with rehearsal. For example, \num{1.7}$\rightarrow$\num{76.1} means the fine-tuned model gets only \num{1.7}\% on Batch 1 tools after CL-Round 2 without rehersal, and \num{76.1}\% with rehersal. While vanilla fine-tuning causes catastrophic forgetting, rehearsal could largely mitigate this issue and allow the model to continually learn new tools while preserving its previously acquired skills.}
\label{tbl:cl}
\vspace{-10pt}
\end{table*}

\subsection{Error Analysis}
\noindent\textbf{Errors of GPT-4.}
As one of the most capable LLMs, GPT-4 (\ttsmall{8k-0613}) can only achieve an overall correctness of \num{60.8}\%.
We conduct an error analysis of GPT-4.
We randomly sample and examine \num{30} error examples of GPT-4, which can be categorized into the following three types, with the corresponding percentage without $\rightarrow$ with ICL with STE examples.
Examples for each category are shown in Appendix~\ref{app:err}.
\begin{itemize}[noitemsep,nolistsep,leftmargin=*]
    \item \textbf{Wrong choice of API (36.7\%$\rightarrow$19.0\%).} GPT-4 calls the wrong API that cannot address the user query. 
    % While it is hard to know the exact underlying reason for such errors, we believe that the cause is a combination of API overlooking and lack of API understanding. 
    Table~\ref{qual_api} shows one example where the user query is regarding parks with hiking trails in San Francisco. Here the model calls an API that retrieves the geographic coordinates of San Francisco, overlooking the ground truth ``Places'' API.
    ICL with STE examples helps resolve about half of such errors by better illustrating the fine-grained semantics of the APIs.
    \item \textbf{Missing/wrong arguments (26.7\%$\rightarrow$10.0\%).} Here, GPT-4 fails to provide the correct set of arguments despite choosing the right tool. Table~\ref{qual_args} shows an example where the model fails to provide the required ``lang'' keyword.
    STE is particularly effective for such errors.
    \item \textbf{Hard-to-evaluate examples (36.7\%$\rightarrow$16.7\%).} We found that it is difficult to judge the correctness of the model predictions for around one-third of the error examples (an example included in Table~\ref{qual_hard}).
    The main reasons behind this are 1)~the existence of tools with overlapping functionalities that makes ground truth non-unique and 2)~the time-sensitive nature of certain tools that prohibits consistent ground truths. 
    Such difficulties in evaluating tool use are also noted in existing work~\cite{qin2023toolllm, patil2023gorilla}, which is an open challenge for future work.
\end{itemize}

\noindent\textbf{Errors after fine-tuning.} We also examine the errors of the most performant fine-tuned model (Mistral-Instruct-7B) and summarize the notable error causes compared with GPT-4, which shed light on venues for future improvement.
\begin{itemize}[noitemsep,nolistsep,leftmargin=*]
    \item \textbf{Commonsense/world knowledge (47.4\%).} Many tools require commonsense/world knowledge. Figure~\ref{fig:mistral}(a) shows an example where calling the API requires knowing the \num{4}-character abbreviation of the target transit station, and here the model hallucinates a wrong abbreviation. This issue could be mitigated by scaling or additional knowledge retrieval.
    \item \textbf{Language understanding (31.6\%).} Certain errors are caused by a lack of basic language understanding abilities. Figure~\ref{fig:mistral}(b) shows one example where the model misunderstands the user query which results in wrong arguments. Using a stronger base LLM could mitigate such errors.
    \item \textbf{Grounding (21.1\%).} We find that some errors are due to a lack of grounding, where the LLM generates API calls that are semantically correct but not grounded to the API constraints. One example is given in Figure~\ref{fig:mistral}(c), where the model correctly extracts the target entity but fails to link it to the entity names supported by the API. This could be improved by incorporating constraints during decoding~\citep{zhang2023syntax,shin-etal-2021-constrained,fang-etal-2023-whole} or using fuzzy-matching mechanisms.

\end{itemize}

\subsection{Continual Tool Learning}
\label{sec:cl}
While fine-tuning significantly outperforms ICL for tool use, one downside is the potential decrease of \textit{flexibility} as discussed in \S\ref{sec:intro} due to catastrophic forgetting~\citep{kirkpatrick2017overcoming, howard-ruder-2018-universal, kumar2022finetuning, luo2023empirical}. Since retraining the model from scratch is costly and hurts flexibility, we explore continual learning (CL) and show that simple rehearsal~\citep{scialom-etal-2022-fine} seems to be sufficient for continual tool learning with STE.

We randomly split the tools into 4 consecutive batches to simulate the continual setting.
For rehearsal, during each round, we add \num{10}\% tool use examples for each API from previous batches into the replay buffer.
For preserving general non-tool-use capabilities, we also add in every training round \num{2000} random examples from Flan-V2~\citep{longpre2023flan, chung2022scaling}, one of the highest quality general instruction datasets~\citep{wang2023how}, and evaluate the model on MMLU~\citep{hendrycks2021measuring} and Big-Bench-Hard (BBH)~\citep{suzgun2022challenging}.
We use Llama-Flan as the base model to ensure a fair comparison of general capabilities on MMLU and BBH (more details in Appendix~\ref{app:other}).
Results in Table~\ref{tbl:cl} show that the model could drastically forget previously learned tools without rehearsal, with more distant ones being more severely forgotten. Rehearsal largely mitigates forgetting---the CL-trained model achieves comparable performance as Llama-FT. General language abilities are also retained as measured on MMLU and BBH. Overall, we extend the findings of \citet{scialom-etal-2022-fine} on the effectiveness of experience replay to the new realm of LLM tool learning, demonstrating a feasible way of flexibly adding new tools with the proposed STE method.
\section{Related Work}

\noindent\textbf{Tool-augmented language models.} One of the focuses of extensive research in NLP is on augmenting models with retrieval/search engines that could supplement extra knowledge~\citep[][\textit{inter alia}]{guu2020retrieval, lewis2020retrieval, izacard2022atlas, borgeaud2022improving}. Recently, there has been a trend towards augmenting LLMs with more diverse types of tools, such as program executors, translation and QA models~\citep{chen2022program, gao2023pal, parisi2022talm, schick2023toolformer}, APIs from developers and public repositories~\citep{patil2023gorilla, qin2023toolllm, xie2023openagents}, and tools curated for specific environments~\cite{gu2024middleware} to further expand the scope of problems that LLMs can assist with.

Both fine-tuning and ICL are used to adapt an LLM to use tools. Fine-tuning-based approaches train the LLM to use tools on a set of tool-specific demonstration examples~\citep{schick2023toolformer, parisi2022talm}, while ICL-based approaches~\citep{lu2023chameleon, song2023restgpt, shen2023hugginggpt, liang2023taskmatrix, gao2023pal} directly put the tool descriptions and optionally (a small amount of) tool-use demonstrations in the context. \citet{hao2023toolkengpt} propose a lightweight adaptation method that expands the LLM's vocabulary with trained tool embeddings. \citet{qin2023toolllm, patil2023gorilla, tang2023toolalpaca} explore training models to better leverage API descriptions for tool use. Our work aims to develop a framework that allows equipping LLMs with stronger tool-use abilities, motivated by how humans typically learn tools through continual trial and error.

\noindent\textbf{LLMs can learn from feedback.} Recent work found that LLMs are capable of improving/correcting their predictions with feedback~\citep{shinn2023reflexion, madaan2023self, ganguli2023capacity, chen2023teaching, peng2023check, kim2023language, pan2023automatically}. Our work is built on top of these findings and uses an LLM to progressively learn tools by leveraging feedback from the tool execution and the LLM's self-reflection.

\noindent\textbf{Data synthesis \& bootstrapping with LLMs.} Due to LLMs' exposure to broad domains during pretraining and their rapidly improving generation abilities, recent work has explored using LLMs for dataset synthesis, which alleviates the burden of costly human annotations~\citep{schick-schutze-2021-generating, wang2022self, honovich-etal-2023-unnatural, li2023selfalignment, zelikman2022star, huang-etal-2023-large}. Such model-synthesized data can then be utilized to improve models including themselves. In the tool-learning domain, similar ideas have been explored for tool-specific data synthesis~\citep{schick2023toolformer, patil2023gorilla, qin2023toolllm}. Our approach follows this line of work and takes a step towards better comprehensiveness and diversity of the synthesized tool-use examples.

\noindent\textbf{Augmenting models with dynamic memory.} Using memory mechanisms to allow models to dynamically gather and utilize experiences is an old idea, e.g., \citet{riesbeck1981failure, 10.5555/538776}. Recent work also explores augmenting models with a growing memory of user and environment feedback~\citep{madaan-etal-2022-memory, shinn2023reflexion, zhong2023memorybank, liang2023unleashing, zhao2023expel, modarressi2023retllm, hu2023chatdb}. We draw inspiration from these works and augment the LLM with fine-grained short-term memory and distilled long-term memory to enhance the LLM's progressive learning of tools.
\section{Conclusions}
Motivated by how humans master tools through continual interaction and reinforcement, we propose simulated trial and error, an LLM tool-learning method built upon progressive memory-based trial and error. Experiments on APIs drawn from ToolBench show the effectiveness of the proposed method, and also that rehearsal-based fine-tuning could enable continual learning of new tools with preserved previous skills.
\section*{Limitations}
\noindent\textbf{Iterative improvement.} Currently, we use strong models for exploration and smaller weak models for exploitation. The exploration-exploitation could also be done iteratively as in prior work~\citep{aksitov2023rest, zelikman2022star}, where the reliance on the strong models could be diminished gradually (e.g., only as evaluators) as the capabilities of the models being enhanced improve.

\noindent\textbf{Compositional tool use \& planning.} Another important ability in the context of tool use is composing/planning multiple tool calls to fulfill complex queries, which goes in an orthogonal direction as our focus here. Recent works show that the core abilities of LLMs are encoded and elicited from pretraining instead of injected through fine-tuning/alignment~\citep{zhou2023lima, lin2023unlocking}, which suggests that extensive data preparation may not be required to adapt LLMs for complex tool use, different from our focus where extensive learning and exploration are always desired as information is gained from the tool side.

\noindent\textbf{Larger memory capacity beyond context limit.} The capacity of the augmented memory is limited by the context length of the LLM. There are different kinds of approaches that could be used to further scale up the memory, such as using additional retrieval modules~\citep{wang-li-2023-learning} or having more hierarchical/compressed representations of the memory~\citep{chen2023walking}.

\noindent\textbf{Tool unlearning?} While we explored continual learning of new tools, the problem of unlearning is also important as tools could get constantly unloaded/outdated. Knowledge unlearning is generally a challenging problem~\citep{si2023knowledge}, and there could be specific designs that support easier tool unlearning, such as ToolkenGPT~\citep{hao2023toolkengpt} which allows plug-and-play adaptation while enabling learning with large-scale examples.

\noindent\textbf{Limitations of example-based fine-tuning.} Finally, there are also inherent limitations of example-based methods for tool learning, in particular, the difficulty of teaching the model when \textit{not} to use a tool through positive tool-use examples alone. Some potential ways of improving this issue are incorporating negative examples (e.g., using contrastive objectives) or carrying such parts of the API alongside example-based training. We leave these investigations to future work.

% Table~\ref{mistral_lim} shows one such example, where the model generalizes from training samples and calls an API upon a query in a similar domain (tourist information), while the API cannot handle the target domain.

% \section*{Acknowledgements}

\bibliography{anthology,custom}

\onecolumn
\pagebreak
\twocolumn
\appendix

% \newpage
% \setcounter{section}{0}
% \renewcommand{\thesection}{\Alph{section}}  % Use letters for appendix sections

% \DoToC

% \clearpage

\section{API Selection}
\label{app:api}
We select 50 APIs from ToolBench\footnote{https://github.com/OpenBMB/ToolBench}, a large collection of APIs from BMTools\footnote{https://github.com/OpenBMB/BMTools} and RapidAPI\footnote{https://rapidapi.com/}. For BMTools, we manually selected 10 high-quality APIs. For RapidAPI, due to the large scale of API pool (over 16k), we perform filtering by 1) selecting APIs that are free to use without subscription needed; 2) sorting the APIs based on the rate limit and latency and selecting the top 400 ones; 3) running a small scale exploration using ChatGPT to interact with the API execution environment, and selecting the top 40 ones based on the API call success rate.

% \section{Full Main Results}
% \label{app:full}
% Table~\ref{tbl:main} includes the main results for all metrics.
% \input{tables/main}

\section{Experimental Details}  
\label{app:other}
\textbf{Fine-tuning.} All model fine-tuning are done using the AdamW optimizer~\citep{loshchilov2018decoupled} with $\beta=(0.9, 0.999)$, $\epsilon=10^{-8}$ and no weight decay. We fine-tune the model for one epoch with learning rate $2\times 10^{-5}$, and a cosine scheduler with warmup ratio \num{0.03}, and batch size \num{64}. All fine-tuning runs are done with \num{4} NVIDIA A100/A6000 GPUs. The training examples are truncated to \num{2048} tokens.

\noindent\textbf{Flan-V2 data \& rehearsal}. We use Flan-V2~\citep{chung2022scaling, longpre2023flan} for rehearsal which is one of the highest quality general-domain instruction tuning data~\citep{wang2023how}. We use the data released by \citet{wang2023how}.\footnote{https://github.com/allenai/open-instruct}

\section{Example queries for all explored trials}
\label{app:examples}
\begin{figure*}[t]
  \centering
    \includegraphics[width=1.0\linewidth]{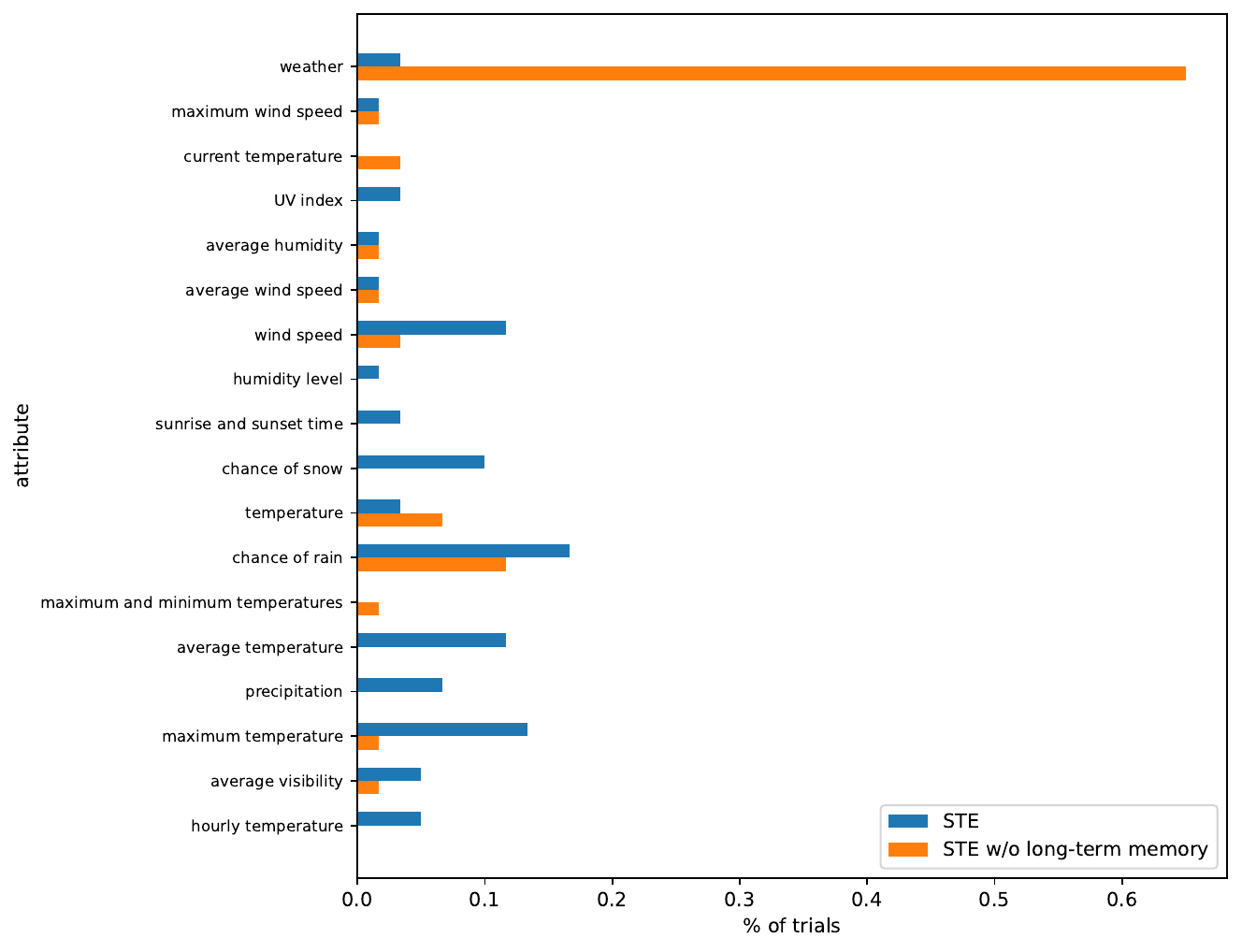}
  \caption{Histogram of explored trials grouped according to the inquired attribute. Long-term memory substantially improves the diversity of the explored attributes.}
\label{fig:hist}
\end{figure*}

Table~\ref{tbl:q}, \ref{tbl:q_no_short}, \ref{tbl:q_no_long} include the queries for all explored trials for the \ttsmall{forecast\_weather} API across different settings (as indicated in the caption). Figure~\ref{fig:hist} includes the histogram of explored trials grouped according to the inquired attribute, comparing STE and STE without long-term memory.
The results clearly show that the memory mechanisms in STE substantially improve the diversity and comprehensiveness of the exploration.

\section{Error Examples of GPT-4 and Mistral-Instruct-7B}
\label{app:err}
Table~\ref{qual_api}, \ref{qual_args}, \ref{qual_hard} include the error examples of GPT-4, and Table~\ref{mistral_know}, \ref{mistral_lu}, \ref{mistral_grounding} include the error examples of fine-tuned Mistral-Instruct-7B referred by \S\ref{sec:analysis}.

\section{Prompts}
\label{sec:prompts}

Table~\ref{tab:exploration}, \ref{tab:longmem} include the full prompt for simulated trial and error. Table~\ref{tab:exploration} includes the prompt for self-refine with execution feedback and short-term memory, and Table~\ref{tab:longmem} includes the prompt for long-term memory. Table~\ref{tab:filtering} and \ref{tab:paraphrasing} includes the prompts for example filtering and paraphrasing respectively.

\vfill
\pagebreak

\begin{table*}[ht]
\centering
\resizebox{1.0\linewidth}{!}{
\begin{tabular}{p{17cm}}
     \toprule
        \textbf{User query}: Which parks in San Francisco have hiking trails? \\
        \textbf{Ground truth API name}: \myblue{search\_places} \\
        \textbf{Description for ground truth API}: Run Places search. Your input should be a json (args json schema): \{"query": string, \} \\
        \textbf{Ground truth arguments}: \{"query": "parks in San Francisco with hiking trails"\} \\
        \hdashline
        \textbf{Predicted API}: \mypink{Geographic coordinates by placename} \\
        \textbf{Description for predicted API}: Returns geographic coordinates for the given placename (city, village, etc.). The method returns the place whose name is most similar to the search string.", "required\_parameters": [\{"name": "name", "type": "STRING", "description": "Placename", "default": "London"\}, \{"name": "lang", "type": "ENUM", "description": "Two-letter language code (ISO639-1). The following values are available: en (english), ru (russian)", "default": ""\}], "optional\_parameters": [\{"name": "country", "type": "STRING", "description": "Two-letter country code, ISO-3166 (optional). Default is all countries.", "default": ""\}] \\
        \textbf{Predicted Arguments}: \{"name": "San Francisco", "lang": "en"\} \\
     \bottomrule
\end{tabular}}
\caption{Error example for GPT-4 (category: wrong API call). Here GPT-4 ``overlooks'' the Google Places API which fulfills the query, and instead calls an API that retrieves the geographic coordinates of the target place.}
\label{qual_api}
\end{table*}

\begin{table*}[ht]
\centering
\resizebox{1.0\linewidth}{!}{
\begin{tabular}{p{17cm}}
     \toprule
        \textbf{User query}: What are the geographic coordinates for the city of Sydney, Canada? \\
        \textbf{Ground truth API name}: Geographic coordinates by placename \\
        \textbf{Description for ground truth API}: Returns geographic coordinates for the given placename (city, village, etc.). The method returns the place whose name is most similar to the search string. Required parameters: [\{"name": "name", "type": "STRING", "description": "Placename", "default": "London"\}, \{"name": "lang", "type": "ENUM", "description": "Two-letter language code (ISO639-1). The following values are available: en (english), ru (russian)", "default": ""\}]. Optional parameters: [\{"name": "country", "type": "STRING", "description": "Two-letter country code, ISO-3166 (optional). Default is all countries.", "default": ""\}] \\
        \textbf{Ground truth arguments}: \myblue{\{"name": "Sydney", "country": "CA", "lang": "en"\}} \\
        \hdashline
        \textbf{Predicted API}: Geographic coordinates by placename \\
        \textbf{Description for predicted API}: Returns geographic coordinates for the given placename (city, village, etc.). The method returns the place whose name is most similar to the search string.", "required\_parameters": [\{"name": "name", "type": "STRING", "description": "Placename", "default": "London"\}, \{"name": "lang", "type": "ENUM", "description": "Two-letter language code (ISO639-1). The following values are available: en (english), ru (russian)", "default": ""\}], "optional\_parameters": [\{"name": "country", "type": "STRING", "description": "Two-letter country code, ISO-3166 (optional). Default is all countries.", "default": ""\}] \\
        \textbf{Predicted Arguments}: \mypink{[\{"name": "Sydney", "country": "CA"\}} \\
     \bottomrule
\end{tabular}}
\caption{Error example for GPT-4 (category: wrong API arguments). Here GPT-4 chooses the right tool but fails to fill in the required ``lang'' keyword.}
\label{qual_args}
\end{table*}

\begin{table*}[ht]
\centering
\resizebox{1.0\linewidth}{!}{
\begin{tabular}{p{17cm}}
     \toprule
        \textbf{User query}: What are the main causes of climate change? \\
        \textbf{Ground truth API name}: \myblue{search} \\
        \textbf{Description for ground truth API}: The input is an exact entity name. The action will search this entity name on Wikipedia and returns the first five sentences if it exists. If not, it will return some related entities to search next. Your input should be a json (args json schema): \{"entity": string, \} \\
        \textbf{Ground truth arguments}: \{"entity": "causes of climate change"\} \\
        \hdashline
        \textbf{Predicted API}: \mypink{search\_general} \\
        \textbf{Description for predicted API}: Run query through GoogleSearch and parse result. Your input should be a json (args json schema): \{"query": string, \} \\
        \textbf{Predicted Arguments}: \{"query": "main causes of climate change"\} \\
     \bottomrule
\end{tabular}}
\caption{Error example for GPT-4 (category: hard to judge correctness). Here GPT-4 calls the Google Search API which could also be a valid choice besides the ground truth.}
\label{qual_hard}
\end{table*}
\begin{table*}[ht]
\centering
\resizebox{1.0\linewidth}{!}{
\begin{tabular}{p{17cm}}
     \toprule
        \textbf{User query}: What is the current advisory information for the \myblue{Union City station}? \\
        \textbf{Ground truth API name}: BART - Advisory information\\
        \textbf{Description for ground truth API}: The BART API gives you access to pretty much all of the BART service and station data available on the BART website. "required\_parameters": [\{"name": "cmd", "type": "STRING", "description": "See more examples http://api.bart.gov/docs/overview/examples.aspx", "default": "bsa"\}], "optional\_parameters": [\{"name": "orig", "type": "STRING", "description": "Optional station filter. Uses 4 character BART station abbreviations (http://api.bart.gov/docs/overview/abbrev.aspx)", "default": ""\}]\\
        \textbf{Ground truth arguments}: \{"cmd": "bsa", \myblue{"orig": "UCTY"}\}\\
        \hdashline
        \textbf{Predicted API}: BART - Advisory information \\
        \textbf{Predicted Arguments}: \{"cmd": "bsa", \mypink{"orig": "UNION"}\}\\
     \bottomrule
\end{tabular}}
\caption{Error example of Mistral-Instruct (7B) after fine-tuning. Here the model hallucinates a wrong abbreviation for the target station due to a lack of commonsense/world knowledge.}
\label{mistral_know}
\end{table*}

\begin{table*}[ht]
\centering
\resizebox{1.0\linewidth}{!}{
\begin{tabular}{p{17cm}}
     \toprule
        \textbf{User query}: What are the ongoing giveaways for PC game keys on the \myblue{GOG platform}? \\
        \textbf{Ground truth API name}: GamerPower - Filter \& Group Giveaways\\
        \textbf{Description for ground truth API}: "Find all free games, loot and giveaways with this giveaway tracker API powered by GamerPower.com! Access programmatically the best giveaways in gaming! Filter and group platforms and giveaway types to get personalized results.", "required\_parameters": [\{"name": "platform", "type": "STRING", "description": "", "default": "epic-games-store.steam.android"\}], "optional\_parameters": [\{"name": "type", "type": "STRING", "description": "", "default": "game.loot"\}] \\
        \textbf{Ground truth arguments}: \{\myblue{"platform": "gog"}, "type": "game.key"\}\\
        \hdashline
        \textbf{Predicted API}: GamerPower - Filter \& Group Giveaways \\
        \textbf{Predicted Arguments}: \{\mypink{"platform": "pc"}, "type": "game.key"\}\\
     \bottomrule
\end{tabular}}
\caption{Error example of Mistral-Instruct (7B) after fine-tuning. Here the model makes a mistake on understanding the user query and predicts the wrong argument for ``platform''.}
\label{mistral_lu}
\end{table*}

\begin{table*}[ht]
\centering
\resizebox{1.0\linewidth}{!}{
\begin{tabular}{p{17cm}}
     \toprule
        \textbf{User query}: Can you provide the total number of matches played between Leeds United and Sheffield Wednesday in the English Premier League? \\
        \textbf{Ground truth API name}: Football Dolphin - Head to head statistics\\
        \textbf{Description for ground truth API}: This Api returns statistical data about English Premier League. Click on the link to view all endpoints in one web app  https://football-dolphin-web-app.up.railway.app/. "required\_parameters": [\{"name": "first\_team", "type": "STRING", "description": "**Enter first team from all available teams:** Arsenal, Aston Villa, Barnsley, Birmingham, Blackburn, Blackpool, Bolton, Bournemouth, Bradford, Brighton, Burnley, Cardiff, Charlton, Chelsea, Coventry, Crystal Palace, Derby, Everton, Fulham, Huddersfield, Hull, Ipswich, Leeds, Leicester, Liverpool, Man City, Man United, Middlesbrough, Newcastle, Norwich, Nott'm Forest, Portsmouth, QPR, Reading, Sheffield United, Sheffield Weds, Southampton, Stoke, Sunderland, Swansea, Tottenham, Watford, West Brom, West Ham, Wigan, Wimbledon, Wolves", "default": "Man United"\}, \{"name": "second\_team", "type": "STRING", "description": "**Enter second team from all available teams:** Arsenal, Aston Villa, Barnsley, Birmingham, Blackburn, Blackpool, Bolton, Bournemouth, Bradford, Brighton, Burnley, Cardiff, Charlton, Chelsea, Coventry, Crystal Palace, Derby, Everton, Fulham, Huddersfield, Hull, Ipswich, \myblue{Leeds}, Leicester, Liverpool, Man City, Man United, Middlesbrough, Newcastle, Norwich, Nott'm Forest, Portsmouth, QPR, Reading, Sheffield United, Sheffield Weds, Southampton, Stoke, Sunderland, Swansea, Tottenham, Watford, West Brom, West Ham, Wigan, Wimbledon, Wolves", "default": "Liverpool"\}, \{"name": "type\_of\_statistics", "type": "STRING", "description": "**Enter one from available types of statistics:** full time result, home vs away full time result, result first half and the match,exact number of goals in the match, goals over, goals under", "default": "full time result"\}] \\
        \textbf{Ground truth arguments}: \{"first\_team": "\myblue{Leeds}", "second\_team": "Sheffield Weds", "type\_of\_statistics": "full time result"\}\\
        \hdashline
        \textbf{Predicted API}: Football Dolphin - Head to head statistics \\
        \textbf{Predicted Arguments}: \{"first\_team": "\mypink{Leeds United}", "second\_team": "Sheffield Weds", "type\_of\_statistics": "full time result"\}\\
     \bottomrule
\end{tabular}}
\caption{Error example of Mistral-Instruct (7B) after fine-tuning. Here the model correctly extracts the target entity but fails to ground it to the set of supported entities of the tool.}
\label{mistral_grounding}
\end{table*}

% \begin{table*}[ht]
% \centering
% \resizebox{1.0\linewidth}{!}{
% \begin{tabular}{p{17cm}}
%      \toprule
%         \textbf{User query}: What are the popular tourist destinations in Bali, Indonesia? \\
%         \textbf{Ground truth API name}: \myblue{search}\\
%         \textbf{Description for ground truth API}: The input is an exact entity name. The action will search this entity name on Wikipedia and returns the first five sentences if it exists. If not, it will return some related entities to search next.. Your input should be a json (args json schema): {"entity" : string}\\
%         \textbf{Ground truth arguments}: \{"entity": "Bali"\}\\
%         \hdashline
%         \textbf{Predicted API}: \mypink{thailand - thai4} \\
%         \textbf{Predicted Arguments}: \{\}\\
%      \bottomrule
% \end{tabular}}
% \caption{Error example of Mistral-Instruct (7B) after fine-tuning.}
% \label{mistral_lim}
% \end{table*}
\begin{table*}[h]
    \centering
    \resizebox{\linewidth}{!}{
    \begin{tabular}{p{20cm}}
       \toprule
         Your task is to answer the user's query as best you can. You have access to the following tools which you can use via API call to help with your response: \\

        \{api\_descriptions\} \\
        
        Now you have the chance to explore the available APIs. You can do this by 1) synthesizing some natural user query that calling the API could help, and 2) trying to respond to the user query with the help of the APIs. Here, you can focus on queries that only require calling the API once. \\
        
        Now, first input your synthesized user query. You should make the query natural - for example, try to avoid using the provided API descriptions or API names in the query, as the user does not know what APIs you have access to. Also, try to make the query as specific as possible. Input just the user query alone; do NOT solve the query for now. \\\\
        
        User Query: \\
        
        =========
        
        Now, try to respond to the query using the available APIs. \\
        
        The format you use the API is by specifying 1) Action: the API function name you'd like to call 2) Action Input: the input parameters of the API call in a json string format. The result of the API call will be returned starting with "Observation:". Remember that you should only perform a SINGLE action at a time, do NOT return a list of multiple actions. \\
        
        Reminder: \\
        1) the only values that should follow "Action:" are: \{api\_names\}\\
        2) use the following json string format for the API arguments:\\
        
        Action Input: \\
        \{ \\
            \ \ \ "key\_1": "value\_1", \\
            \ \ \ ... \\
            \ \ \ "key\_n": "value\_n" \\
        \} \\
        
        Remember to ALWAYS use the following format:
        
        Thought: you should always think about what to do next\\
        Action: the API function name\\
        Action Input: the input parameters of the API call in json string format\\
        Observation: the return result of the API call. This is what I will provide you with; you do not need to repeat it in your response.\\
        ... (this Thought/Action/Action Input/Observation can repeat N times)\\
        Thought: I now know the final answer\\
        Final Answer: the response to the user query\\\\
        
        Begin! Remember that your response should never start with "Observation:" since that is what I will provide you with. Once you have enough information, please immediately use \\Thought: I now know the final answer\\Final Answer:\\\\
        
        User Query (the same you just synthesized): \{query\}\\
        
        =========

        Do you think you successfully answered this query in the end? Respond with "Yes" or "No". \\
        =========
        
        Now you know a bit more about the API. You can synthesize another user query to explore the API a bit further and consolidate your understanding of the API, based on things that you discovered about this API. Again, just input the user query alone; do NOT solve the query for now.\\\\
        
        User Query: \\
        
        =========
        
        Now try to solve the query using the API. Remember to follow the same format, i.e,\\Thought:\\Action:\\Action Input:\\Observation:\\Final Answer:\\        
    \bottomrule
    \end{tabular}}
    \caption{Prompt for self-refine with execution feedback and short-term memory.}
    \label{tab:exploration}
\end{table*}

\begin{table*}[h]
    \centering
    \resizebox{\linewidth}{!}{
    \begin{tabular}{p{20cm}}
    \toprule
    Below are queries you have already explored and whether you successfully solved them with the API's help: \\\\
    \{long\_term\_memory\} \\\\
    Based on these, try to explore queries that can help you understand the API further; avoid synthesizing queries that are too close to the existing ones.\\
    \bottomrule
    \end{tabular}}
    \caption{Prompt for long-term memory.}
    \label{tab:longmem}
\end{table*}

\begin{table*}[h]
    \centering
    \resizebox{\linewidth}{!}{
    \begin{tabular}{p{20cm}}
    \toprule
    An assistant is trying to respond to the user query with the help of some APIs. The APIs that the assistant has access to are as follows:\\
    
    \{api\_descriptions\}\\
    
    Now, your task is to evaluate how well the assistant did the job. Check carefully the following aspects of the assistant's response:\\
    
    1) whether the response answers the user's query in an informative way. For example, if the API calls are unsuccessful and the agent can't find the answer to the request, you should say "No."\\
    2) whether the response is faithful with respect to the execution results of the API calls. The response should not include information that cannot be supported by the API call feedback,\\
    3) whether the assistant used the API calls appropriately. For example, the assistant should always use relevant API calls for queries about up-to-date information or complex calculations,\\
    
    For each of the three aspects, you should say "Yes" or "No" indicating whether the assistant did a good job in that aspect, and explain the reason behind your judgment. Your output should follow the format below, where "<explanation>" should be your actual explanation for the corresponding judgment:\\
    
    1) Yes/No. <explanation>\\
    2) Yes/No. <explanation>\\
    3) Yes/No. <explanation>\\
    
    Now, the user query is: \\
    
    \{query\}\\
    
    The assistant's API calls and the corresponding execution results are:\\
    
    \{chains\}
    
    The assistant's final response is: \\
    \{final\_ans\}
    
    Now, your evaluation is (remember to follow the previous format): \\
    
    \bottomrule
    \end{tabular}}
    \caption{Prompt for example filtering.}
    \label{tab:filtering}
\end{table*}

\begin{table*}[h]
    \centering
    \resizebox{\linewidth}{!}{
    \begin{tabular}{p{20cm}}
    \toprule
    Below you will be given a user query. Try to paraphrase it in a different way while preserving its meaning. The query is:\\
    \{query\}\\
    Your paraphrase of the query: \\
    =========
    
    Can you try to paraphrase it again in a new way? Avoid coming up with something too close to your previous ones. Your paraphrase:\\
    \bottomrule
    \end{tabular}}
    \caption{Prompt for query paraphrasing.}
    \label{tab:paraphrasing}
\end{table*}
\begin{table*}[h]
    \centering
    \resizebox{\linewidth}{!}{
    \begin{tabular}{p{20cm}}
       \toprule
        What is the UV index in San Francisco for the next 3 days? \\
        What is the UV index in Sydney for the next 3 days? \\
        What is the average humidity in Miami Beach for the next 5 days? \\
        What is the average temperature in Los Angeles for the next 3 days? \\
        What is the average temperature in Miami for the next 5 days? \\
        What is the average temperature in San Francisco for the next 3 days? \\
        What is the average temperature in Sydney for the next 10 days? \\
        What is the average visibility in New York City for the next 3 days? \\
        What is the average visibility in Tokyo for the next 4 days? \\
        What is the average wind speed in Chicago for the next 7 days? \\
        What is the chance of rain in London for the next 3 days? \\
        What is the chance of rain in Los Angeles for the next 7 days? \\
        What is the chance of rain in Miami Beach for the next 3 days? \\
        What is the chance of rain in San Francisco for the next 3 days? \\
        What is the chance of rain in San Francisco for the next 5 days? \\
        What is the chance of rain in Seattle for the next 7 days? \\
        What is the chance of rain in Sydney for the next 5 days? \\
        What is the chance of rain in Tokyo for the next 7 days? \\
        What is the chance of snow in Boston tomorrow? \\
        What is the chance of snow in Chicago for the next 5 days? \\
        What is the chance of snow in Denver for the next 3 days? \\
        What is the chance of snow in New York City for the next 5 days? \\
        What is the chance of snow in Paris for the next 3 days? \\
        What is the forecasted wind speed in London for the next 5 days? \\
        What is the hourly precipitation forecast in New York City for the next 24 hours? \\
        What is the hourly precipitation forecast in Seattle for the next 24 hours? \\
        What is the hourly temperature forecast for Chicago tomorrow? \\
        What is the hourly temperature forecast in Los Angeles for the next 24 hours? \\
        What is the hourly temperature forecast in New York City for the next 12 hours? \\
        What is the hourly temperature forecast in San Francisco for the next 12 hours? \\
        What is the hourly wind speed forecast in Miami tomorrow? \\
        What is the maximum temperature in Sydney for the next 7 days? \\
        What is the maximum wind speed in Los Angeles for the next 2 days? \\
        What is the sunrise and sunset time in Paris tomorrow? \\
        What is the sunrise and sunset time in Tokyo tomorrow? \\
        What is the temperature range in Sydney for the next 3 days? \\
        What is the total precipitation in Sydney for the next 10 days? \\
        What is the total precipitation in Tokyo for the next 7 days? \\
        What is the weather forecast in Miami for the next week? \\
        What is the wind forecast in San Francisco for the next 7 days? \\
        What will be the average temperature in London for the next 5 days? \\
        What will be the average temperature in Seattle for the next 7 days? \\
        What will be the average temperature in Tokyo for the next 7 days? \\
        What will be the average visibility in Miami for the next 7 days? \\
        What will be the average wind speed in San Francisco for the next 3 days? \\
        What will be the chance of rain in San Diego for the next 7 days? \\
        What will be the humidity level in London for the next 5 days? \\
        What will be the maximum temperature in Chicago for the next 5 days? \\
        What will be the maximum temperature in London for the next 10 days? \\
        What will be the maximum temperature in Los Angeles for the next 7 days? \\
        What will be the maximum temperature in Madrid for the next 7 days? \\
        What will be the maximum temperature in Paris for the next 7 days? \\
        What will be the maximum temperature in Sydney for the next 5 days? \\
        What will be the maximum temperature in Tokyo for the next 10 days? \\
        What will be the maximum wind speed in San Francisco for the next 5 days? \\
        What will be the weather like in New York City for the next 5 days? \\
        What will be the wind speed forecast in Boston for the next 3 days? \\
        What will be the wind speed in Chicago tomorrow morning? \\
        Will it rain in Seattle tomorrow? \\
        Will it snow in Denver next week? \\
       \bottomrule
    \end{tabular}}
    \caption{Queries in all explored trials (prefix-sorted). Setting: STE (both short-term and long-term memory).}
    \label{tbl:q}
\end{table*}

\begin{table*}[h]
    \centering
    \resizebox{\linewidth}{!}{
    \begin{tabular}{p{20cm}}
       \toprule
        Is it going to rain in London tomorrow? \\
        What will be the temperature in San Francisco on Thursday? \\
        What will be the weather like in Barcelona on Friday? \\
        What will be the weather like in Berlin next Wednesday? \\
        What will be the weather like in Chicago in the next 7 days? \\
        What will be the weather like in Denver in the next 5 days? \\
        What will be the weather like in London in the next 2 days? \\
        What will be the weather like in London in the next 3 days? \\
        What will be the weather like in London in the next 5 days? \\
        What will be the weather like in London next Friday? \\
        What will be the weather like in London on Sunday? \\
        What will be the weather like in Los Angeles in the next 3 days? \\
        What will be the weather like in Los Angeles on Monday? \\
        What will be the weather like in Madrid in the next 5 days? \\
        What will be the weather like in Madrid on Sunday? \\
        What will be the weather like in Miami in the next 7 days? \\
        What will be the weather like in Miami on Friday? \\
        What will be the weather like in Miami on Sunday? \\
        What will be the weather like in Miami on Wednesday? \\
        What will be the weather like in New York City in the next 5 days? \\
        What will be the weather like in New York City next Saturday? \\
        What will be the weather like in New York City on Thursday? \\
        What will be the weather like in Paris in the next 3 days? \\
        What will be the weather like in Paris next Monday? \\
        What will be the weather like in Paris on Sunday? \\
        What will be the weather like in San Antonio in the next 10 days? \\
        What will be the weather like in San Francisco on Friday? \\
        What will be the weather like in San Francisco on Saturday? \\
        What will be the weather like in Seattle in the next 5 days? \\
        What will be the weather like in Sydney in the next 5 days? \\
        What will be the weather like in Sydney in the next 7 days? \\
        What will be the weather like in Sydney next Monday? \\
        What will be the weather like in Sydney next Tuesday? \\
        What will be the weather like in Sydney next Wednesday? \\
        What will be the weather like in Sydney on Monday? \\
        What will be the weather like in Sydney on Saturday? \\
        What will be the weather like in Tokyo in the next 3 days? \\
        What will be the weather like in Tokyo in the next 4 days? \\
        What will be the weather like in Tokyo in the next 7 days? \\
        What will be the weather like in Tokyo next Sunday? \\
        What will be the weather like in Tokyo next Tuesday? \\
        What will be the weather like in Tokyo on Friday? \\
        What will be the weather like in Tokyo on Monday? \\
        What will be the weather like in Tokyo on Saturday? \\
        What will be the weather like in Tokyo on Sunday? \\
        What will be the weather like in Tokyo on Thursday? \\
        What will be the weather like in Tokyo on Wednesday? \\
        What will the weather be like in Barcelona in the next 7 days? \\
        What will the weather be like in Miami next Monday? \\
        What will the weather be like in Paris on Monday? \\
        What will the weather be like in Sydney next Wednesday? \\
        What will the weather be like in Tokyo tomorrow? \\
        Will it be cloudy in Tokyo tomorrow? \\
        Will it be hot in Miami next week? \\
        Will it be rainy in Seattle tomorrow? \\
        Will it be sunny in Madrid tomorrow? \\
        Will it be sunny in San Francisco next Thursday? \\
        Will it be sunny in San Francisco on Wednesday? \\
        Will it rain in London in the next 3 days? \\
        Will it rain in Sydney tomorrow? \\
       \bottomrule
    \end{tabular}}
    \caption{Queries in all explored trials (prefix-sorted). Setting: STE with no short-term memory.}
    \label{tbl:q_no_short}
\end{table*}

\begin{table*}[h]
    \centering
    \resizebox{\linewidth}{!}{
    \begin{tabular}{p{20cm}}
       \toprule
        Can I get the hourly weather forecast for tomorrow in New York City? \\
        Can you provide the weather forecast for the next 10 days in Los Angeles? \\
        How accurate is the weather forecast for the upcoming week? \\
        What is the average visibility in Paris for the next 7 days? \\
        What is the chance of rain in Los Angeles tomorrow? \\
        What is the chance of rain in Miami for the next 3 days? \\
        What is the chance of rain in Seattle tomorrow? \\
        What is the chance of rain in Sydney for the next 7 days? \\
        What is the chance of rain in Tokyo for the next 5 days? \\
        What is the current temperature in Los Angeles? \\
        What is the current temperature in New York City? \\
        What is the current weather condition in Sydney? \\
        What is the forecasted temperature for Paris tomorrow? \\
        What is the forecasted temperature in Los Angeles for the next 10 days? \\
        What is the forecasted weather for New York City for the next 5 days? \\
        What is the forecasted wind speed for Los Angeles tomorrow? \\
        What is the hourly weather forecast for Los Angeles tomorrow? \\
        What is the hourly weather forecast for Los Angeles tomorrow? \\
        What is the weather forecast for London tomorrow? \\
        What is the weather forecast for Los Angeles tomorrow? \\
        What is the weather forecast for Tokyo in the next 10 days? \\
        What is the weather forecast for the next 10 days in London? \\
        What is the weather forecast for the next 2 days in London? \\
        What is the weather forecast for the next 3 days in Paris? \\
        What is the weather forecast for the next 5 days in Los Angeles? \\
        What is the weather forecast for tomorrow in New York City? \\
        What is the wind speed in Paris tomorrow? \\
        What will be the average humidity in Tokyo for the next 3 days? \\
        What will be the average wind speed in Seattle for the next 3 days? \\
        What will be the maximum and minimum temperatures in London tomorrow? \\
        What will be the maximum temperature in London tomorrow? \\
        What will be the maximum wind speed in Paris tomorrow? \\
        What will be the temperature in London in the next 3 days? \\
        What will be the temperature in San Francisco tomorrow? \\
        What will be the weather conditions in London for the next 7 days? \\
        What will be the weather forecast for London tomorrow? \\
        What will be the weather forecast for New York City in the next 5 days? \\
        What will be the weather forecast for Seattle over the next 5 days? \\
        What will be the weather forecast for Sydney next week? \\
        What will be the weather forecast for the next 5 days in New York City? \\
        What will be the weather in New York City for the next 5 days? \\
        What will be the weather like in London for the next 3 days? \\
        What will be the weather like in London for the next 7 days? \\
        What will be the weather like in New York City for the next 5 days? \\
        What will be the weather like in New York City for the next 5 days? \\
        What will be the weather like in New York City for the next 5 days? \\
        What will be the weather like in New York City for the next three days? \\
        What will be the weather like in New York City in the next 5 days? \\
        What will be the weather like in Paris for the next 10 days? \\
        What will be the weather like in San Francisco for the next 5 days? \\
        What will be the weather like in San Francisco tomorrow? \\
        What will be the weather like in Tokyo for the next 7 days? \\
        What will the weather be like in London for the next 5 days? \\
        What will the weather be like in London for the next 7 days? \\
        What will the weather be like in Los Angeles for the next 3 days? \\
        What will the weather be like in New York City for the next 5 days? \\
        What will the weather be like in New York City for the next 5 days? \\
        What will the weather be like in New York City tomorrow? \\
        Will it rain in Los Angeles tomorrow? \\
        Will it rain in San Francisco tomorrow? \\

       \bottomrule
    \end{tabular}}
    \caption{Queries in all explored trials (prefix-sorted). Setting: STE with no long-term memory.}
    \label{tbl:q_no_long}
\end{table*}

\end{document}